\documentclass[10pt,twocolumn,letterpaper]{article}

\usepackage{cvpr}
\usepackage{times}
\usepackage{epsfig}
\usepackage{graphicx}
\usepackage{amsmath}
\usepackage{amssymb}
\usepackage[misc]{ifsym}
\usepackage[linesnumbered,ruled]{algorithm2e}

\usepackage[x11names,dvipsnames,table]{xcolor} 
\usepackage{multirow}
\usepackage{array}
\newcolumntype{L}[1]{>{\raggedright\let\newline\\\arraybackslash\hspace{0pt}}m{#1}}
\newcolumntype{C}[1]{>{\centering\let\newline\\\arraybackslash\hspace{0pt}}m{#1}}
\newcolumntype{R}[1]{>{\raggedleft\let\newline\\\arraybackslash\hspace{0pt}}m{#1}}

\usepackage[breaklinks=true,bookmarks=false]{hyperref}

\cvprfinalcopy 

\def\cvprPaperID{****} 

\begin{document}

\title{Recurrent Saliency Transformation Network: \\ Incorporating Multi-Stage Visual Cues for Small Organ Segmentation}

\author{Qihang Yu\textsuperscript{1}, Lingxi Xie\textsuperscript{2}$^{(\textrm{\Letter})}$, Yan Wang\textsuperscript{2},
Yuyin Zhou\textsuperscript{2}, Elliot K. Fishman\textsuperscript{3}, Alan L. Yuille\textsuperscript{2}\\
\textsuperscript{1}Peking University\quad\textsuperscript{2}The Johns Hopkins University\quad
\textsuperscript{3}The Johns Hopkins Medical Institute\\
{\tt\small \{yucornetto,198808xc,wyanny.9,zhouyuyiner,alan.l.yuille\}@gmail.com}\quad
{\tt\small efishman@jhmi.edu}\\
}

\maketitle

\begin{abstract}
We aim at segmenting small organs (e.g., the pancreas) from abdominal CT scans.
As the target often occupies a relatively small region in the input image,
deep neural networks can be easily confused by the complex and variable background.
To alleviate this, researchers proposed a coarse-to-fine approach~\cite{Zhou_2017_Fixed},
which used prediction from the first (coarse) stage to indicate a smaller input region for the second (fine) stage.
Despite its effectiveness, this algorithm dealt with two stages individually,
which lacked optimizing a global energy function, and limited its ability to incorporate multi-stage visual cues.
Missing contextual information led to unsatisfying convergence in iterations,
and that the fine stage sometimes produced even lower segmentation accuracy than the coarse stage.

This paper presents a {\bf Recurrent Saliency Transformation Network}.
The key innovation is a saliency transformation module,
which repeatedly converts the segmentation probability map from the previous iteration as spatial weights
and applies these weights to the current iteration.
This brings us two-fold benefits.
In training, it allows joint optimization over the deep networks dealing with different input scales.
In testing, it propagates multi-stage visual information throughout iterations to improve segmentation accuracy.
Experiments in the NIH pancreas segmentation dataset demonstrate the state-of-the-art accuracy,
which outperforms the previous best by an average of over $2\%$.
Much higher accuracies are also reported on several small organs in a larger dataset collected by ourselves.
In addition, our approach enjoys better convergence properties, making it more efficient and reliable in practice.
\end{abstract}

\section{Introduction}
\label{Introduction}

This paper focuses on small organ ({\em e.g.}, the {\em pancreas}) segmentation from abdominal CT scans,
which is an important prerequisite for enabling computers to assist human doctors for clinical purposes.
This problem falls into the research area named {\em medical imaging analysis}.
Recently, great progress has been brought to this field by the fast development of deep learning,
especially convolutional neural networks~\cite{Krizhevsky_2012_ImageNet}\cite{Long_2015_Fully}.
Many conventional methods, such as the graph-based segmentation approaches~\cite{Ali_2007_Graph}
or those based on handcrafted local features~\cite{Wang_2014_Geodesic},
have been replaced by deep segmentation networks,
which typically produce higher segmentation accuracy~\cite{Ronneberger_2015_UNet}\cite{Roth_2015_DeepOrgan}.

\newcommand{\figurewidth}{7.0cm}
\begin{figure}[t]
\begin{center}
    \includegraphics[width=\figurewidth]{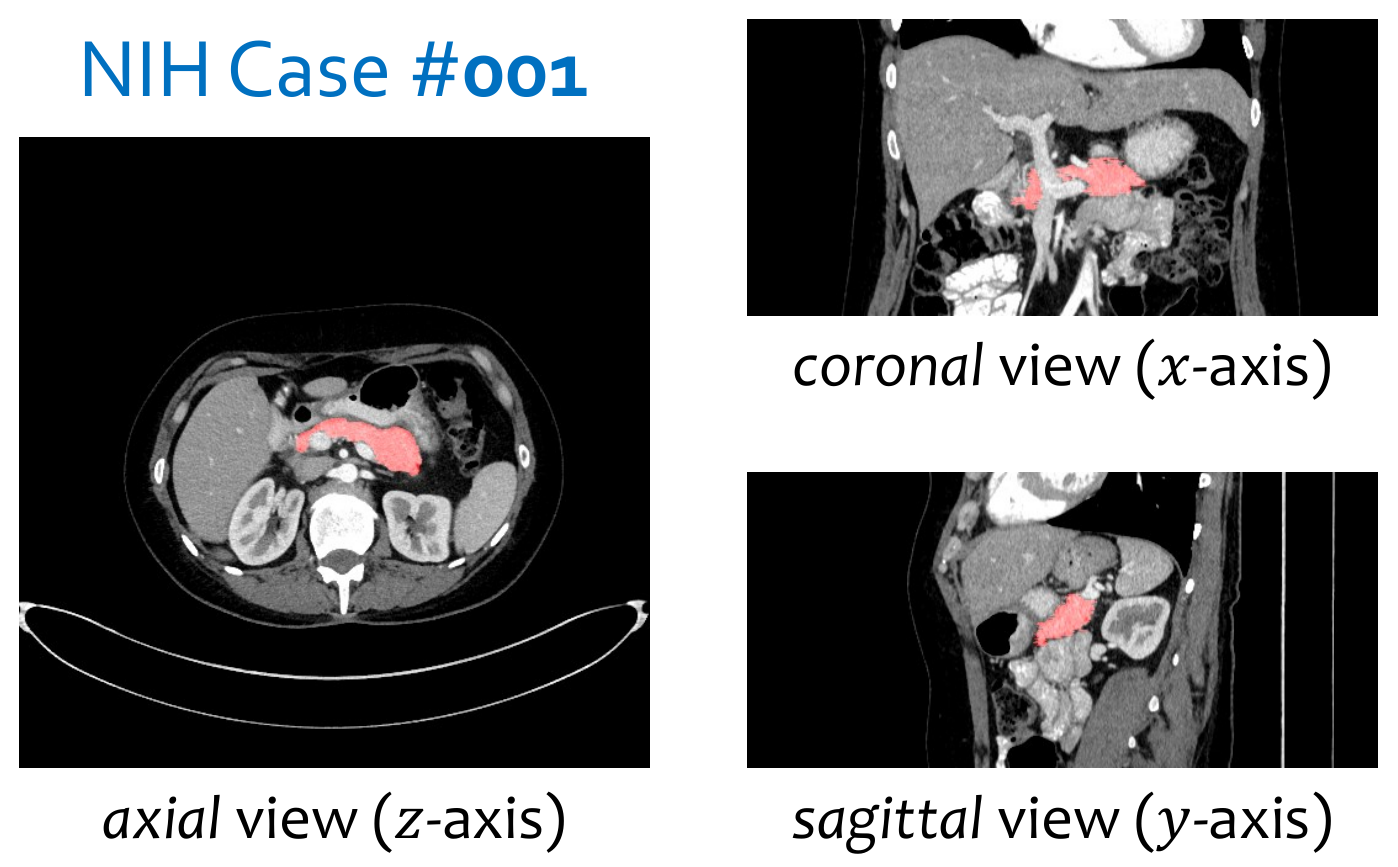}
\end{center}
\caption{
    A typical example from the NIH {\em pancreas} segmentation dataset~\cite{Roth_2015_DeepOrgan} (best viewed in color).
    We highlight the {\em pancreas} in red seen from three different viewpoints.
    It is a relatively small organ with irregular shape and boundary.
}
\label{Fig:Dataset}
\end{figure}

Segmenting a small organ from CT scans is often challenging.
As the target often occupies a {\em small part} of input data
({\em e.g.}, less than $1.5\%$ in a 2D image, see Figure~\ref{Fig:Dataset}),
deep segmentation networks such as FCN~\cite{Long_2015_Fully} and DeepLab~\cite{Chen_2015_Semantic}
can be easily confused by the background region, which may contain complicated and variable contents.
This motivates researchers to propose a {\em coarse-to-fine} approach~\cite{Zhou_2017_Fixed} with two {\em stages},
in which the coarse stage provides a rough localization and the fine stage performs accurate segmentation.
But, despite state-of-the-art performance achieved in pancreas segmentation,
this method suffers from {\em inconsistency} between its training and testing flowcharts,
which is to say, the training phase dealt with coarse and fine stages individually and did not minimize a global energy function,
but the testing phase assumed that these two stages can cooperate with each other in an iterative process.
From another perspective, this also makes it difficult for multi-stage visual cues to be incorporated in segmentation,
{\em e.g.}, the previous segmentation mask which carries rich information is discarded except for the bounding box.
As a part of its consequences, the fine stage consisting of a sequence of iterations cannot converge very well,
and sometimes the fine stage produced even lower segmentation accuracy than the coarse stage (see Section~\ref{Approach:Baseline}).

Motivated to alleviate these shortcomings, we propose a {\bf Recurrent Saliency Transformation Network}.
The chief innovation is to relate the coarse and fine stages with a saliency transformation module,
which repeatedly transforms the segmentation probability map from previous iterations as spatial priors in the current iteration.
This brings us two-fold advantages over~\cite{Zhou_2017_Fixed}.
First, in the training phase, the coarse-scaled and fine-scaled networks are optimized jointly,
so that the segmentation ability of each of them gets improved.
Second, in the testing phase, the segmentation mask of each iteration is preserved and propagated throughout iterations,
enabling multi-stage visual cues to be incorporated towards more accurate segmentation.
To the best of our knowledge, this idea was not studied in the computer vision community,
as it requires making use of some special properties of CT scans (see Section~\ref{Approach:Discussions}).

We perform experiments on two CT datasets for small organ segmentation.
On the NIH {\em pancreas} segmentation dataset~\cite{Roth_2015_DeepOrgan},
our approach outperforms the state-of-the-art by an average of over $2\%$,
measured by the average Dice-S{\o}rensen coefficient (DSC).
On another multi-organ dataset collected by the radiologists in our team,
we also show the superiority of our approach over the baseline on a variety of small organs.
In the testing phase, our approach enjoys better convergence properties,
which guarantees its efficiency and reliability in real clinical applications.

The remainder of this paper is organized as follows.
Section~\ref{RelatedWork} briefly reviews related work, and Section~\ref{Approach} describes the proposed approach.
After experiments are shown in Sections~\ref{ExperimentsNIH} and~\ref{ExperimentsMutliOrgan},
we draw our conclusions in Section~\ref{Conclusions}.

\section{Related Work}
\label{RelatedWork}

Computer-aided diagnosis (CAD) is an important technique which can assist human doctors in many clinical scenarios.
An important prerequisite of CAD is medical imaging analysis.
As a popular and cheap way of medical imaging,
contrast-enhanced computed tomography (CECT) produces detailed images of internal organs, bones, soft tissues and blood vessels.
It is of great value to automatically segment organs and/or soft tissues from these CT volumes
for further diagnosis~\cite{Brosch_2016_Deep}\cite{Wang_2016_Deep}\cite{Havaei_2017_Brain}\cite{Zhou_2017_Deep}.
To capture specific properties of different organs, researchers often design individualized algorithms for each of them.
Typical examples include the the liver~\cite{Ling_2008_Hierarchical}\cite{Heimann_2009_Comparison},
the {\em spleen}~\cite{Linguraru_2010_Automated}, the {\em kidneys}~\cite{Lin_2006_Computer}\cite{Ali_2007_Graph},
the {\em lungs}~\cite{Hu_2001_Automatic}, the {\em pancreas}~\cite{Chu_2013_Multi}\cite{Wang_2014_Geodesic}, {\em etc}.
Small organs ({\em e.g.}, the {\em pancreas}) are often more difficult to segment,
partly due to their low contrast and large anatomical variability in size and (most often irregular) shape.

Compared to the papers cited above which used conventional approaches for segmentation,
the progress of deep learning brought more powerful and efficient solutions.
In particular, convolutional neural networks have been widely applied to a wide range of vision tasks,
such as image classification~\cite{Krizhevsky_2012_ImageNet}\cite{Simonyan_2015_Very}\cite{He_2016_Deep},
object detection~\cite{Girshick_2014_Rich}\cite{Ren_2015_Faster},
and semantic segmentation~\cite{Long_2015_Fully}\cite{Chen_2015_Semantic}.
Recurrent neural networks, as a related class of networks,
were first designed to process sequential data~\cite{Graves_2013_Speech}\cite{Socher_2011_Parsing},
and later generalized to image classification~\cite{Liang_2015_Recurrent} and scene labeling~\cite{Pinheiro_2014_Recurrent} tasks.
In the area of medical imaging analysis, in particular organ segmentation,
these techniques have been shown to significantly outperform conventional approaches,
{\em e.g.}, segmenting the {\em liver}~\cite{Dou_2016_3D}, the {\em lung}~\cite{Harrison_2017_Progressive},
or the {\em pancreas}~\cite{Roth_2016_Spatial}\cite{Cai_2017_Improving}\cite{Roth_2017_Spatial}.
Note that medical images differ from natural images in that data appear in a volumetric form.
To deal with these data, researchers either slice a 3D volume into 2D slices (as in this work),
or train a 3D network
directly~\cite{Merkow_2016_Dense}\cite{Milletari_2016_VNet}\cite{Kamnitsas_2017_Efficient}\cite{Yu_2017_Volumetric}.
In the latter case, limited GPU memory often leads to patch-based training and testing strategies.
The tradeoff between 2D and 3D approaches is discussed in~\cite{Lai_2015_Deep}.

By comparison to the entire CT volume, the organs considered in this paper often occupy a relatively small area.
As deep segmentation networks such as FCN~\cite{Long_2015_Fully} are less accurate in depicting small targets,
researchers proposed two types of ideas to improve detection and/or segmentation performance.
The first type involved rescaling the image so that the target becomes comparable to the training samples~\cite{Xia_2016_Zoom},
and the second one considered to focus on a subregion of the image for each target
to obtain higher accuracy in detection~\cite{Chen_2016_Mitosis} or segmentation~\cite{Zhou_2017_Fixed}.
The coarse-to-fine idea was also well studied in the computer vision area
for saliency detection~\cite{Kuen_2016_Recurrent} or semantic segmentation~\cite{Li_2017_Instance}\cite{Lin_2017_RefineNet}.
This paper is based on a recent coarse-to-fine framework~\cite{Zhou_2017_Fixed},
but we go one step further by incorporating multi-stage visual cues in optimization.

\vspace{-0.02cm}
\section{Our Approach}
\label{Approach}

We investigate the problem of segmenting an organ from abdominal CT scans.
Let a CT image be a 3D volume $\mathbf{X}$ of size $W\times H\times L$
which is annotated with a binary ground-truth segmentation $\mathbf{Y}$ where ${y_i}={1}$ indicates a foreground voxel.
The goal of our work is to produce a binary output volume $\mathbf{Z}$ of the same dimension.
Denote $\mathcal{Y}$ and $\mathcal{Z}$ as the set of foreground voxels in the ground-truth and prediction,
{\em i.e.}, ${\mathcal{Y}}={\left\{i\mid y_i=1\right\}}$ and ${\mathcal{Z}}={\left\{i\mid z_i=1\right\}}$.
The accuracy of segmentation is evaluated by the Dice-S{\o}rensen coefficient (DSC):
${\mathrm{DSC}\!\left(\mathcal{Y},\mathcal{Z}\right)}=
    {\frac{2\times\left|\mathcal{Y}\cap\mathcal{Z}\right|}{\left|\mathcal{Y}\right|+\left|\mathcal{Z}\right|}}$.
This metric falls in the range of $\left[0,1\right]$ with $1$ implying perfect segmentation.

\subsection{Coarse-to-Fine Segmentation and Drawbacks}
\label{Approach:Baseline}

\renewcommand{\figurewidth}{8.0cm}
\begin{figure}[t]
\begin{center}
    \includegraphics[width=\figurewidth]{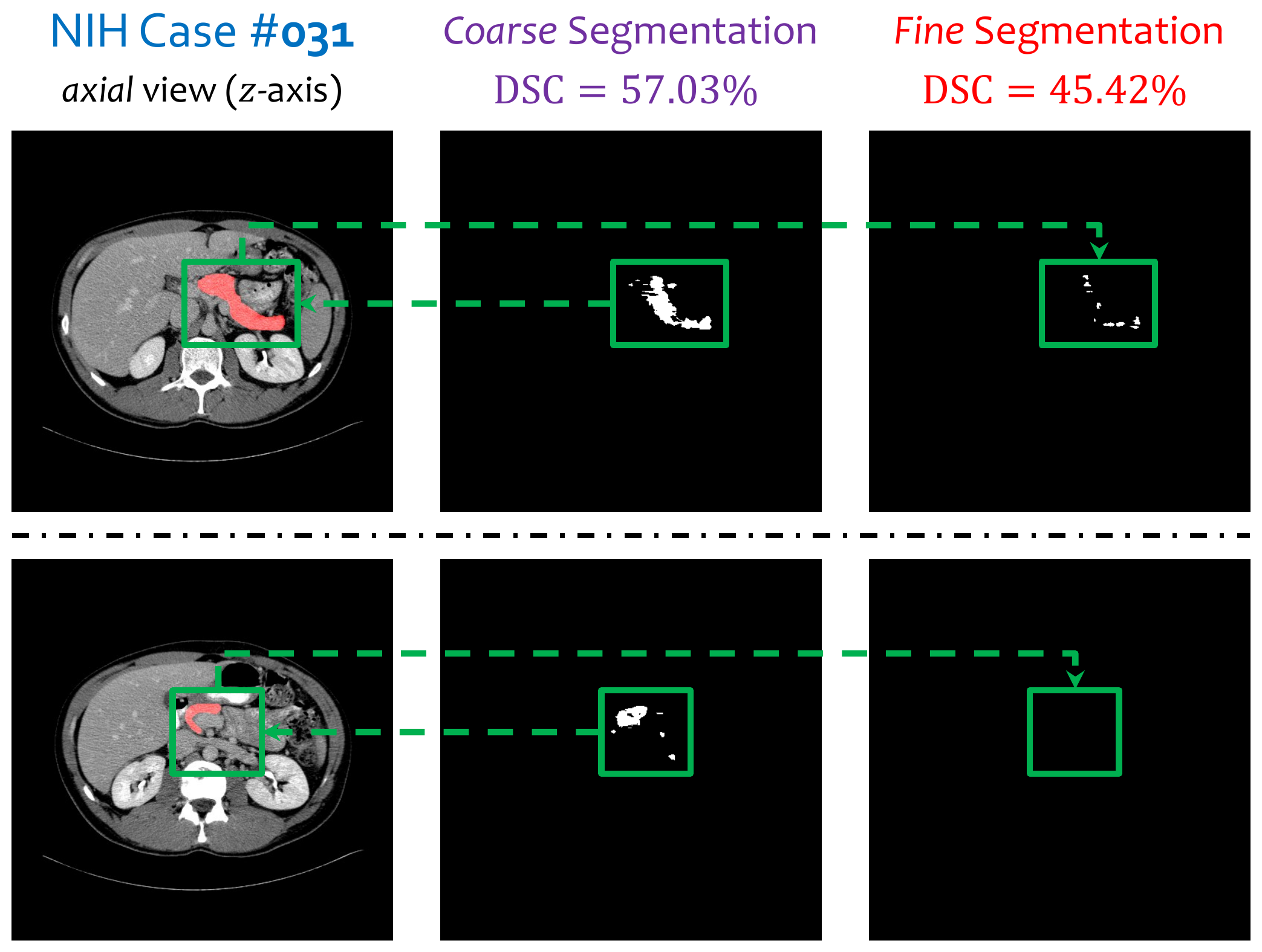}
\end{center}
\caption{
    A failure case of the stage-wise {\em pancreas} segmentation approach~\cite{Zhou_2017_Fixed}
    (in the {\em axial} view, best viewed in color).
    The red masks show ground-truth segmentations,
    and the green frames indicate the bounding box derived from the coarse stage.
    In either slice, unsatisfying segmentation is produced at the fine stage,
    because the cropped region does not contain enough contextual information,
    whereas the coarse-scaled probability map carrying such information is discarded.
    This is improved by the proposed Recurrent Saliency Transformation Network, see Figure~\ref{Fig:VisualizationNIH}.
}
\label{Fig:Motivation}
\end{figure}

We start with training 2D deep networks for 3D segmentation\footnote{
Please see Section~\ref{ExperimentsNIH:Comparison} for the comparison to 3D networks.}.
Each 3D volume $\mathbf{X}$ is sliced along three axes, the {\em coronal}, {\em sagittal} and {\em axial} views,
and these 2D slices are denoted by $\mathbf{X}_{\mathrm{C},w}$ (${w}={1,2,\ldots,W}$),
$\mathbf{X}_{\mathrm{S},h}$ (${h}={1,2,\ldots,H}$) and $\mathbf{X}_{\mathrm{A},l}$ (${l}={1,2,\ldots,L}$),
where the subscripts $\mathrm{C}$, $\mathrm{S}$ and $\mathrm{A}$
stand for {\em coronal}, {\em sagittal} and {\em axial}, respectively.
On each axis, an individual 2D-FCN~\cite{Long_2015_Fully} on a $16$-layer VGGNet~\cite{Simonyan_2015_Very} is trained\footnote{
This is a simple segmentation baseline with a relatively shallow network.
Deeper network structures such as ResNet~\cite{He_2016_Deep}
and more complicated segmentation frameworks such as DeepLab~\cite{Chen_2015_Semantic},
while requiring a larger memory and preventing us from training two stages jointly (see Section~\ref{Approach:Formulation}),
often result in lower segmentation accuracy as these models seem to over-fit in these CT datasets.}.
Three FCN models are denoted by $\mathbb{M}_\mathrm{C}$, $\mathbb{M}_\mathrm{S}$ and $\mathbb{M}_\mathrm{A}$, respectively.
We use the DSC loss~\cite{Milletari_2016_VNet} in the training phase
so as to prevent the models from being biased towards the background class.
Both multi-slice segmentation ($3$ neighboring slices are combined as a basic unit in training and testing)
and multi-axis fusion (majority voting over three axes) are performed to incorporate pseudo-3D information into segmentation.

The organs investigated in this paper ({\em e.g.}, the {\em pancreas}) are relatively small.
In each 2D slice, the fraction of the foreground pixels is often smaller than $1.5\%$.
To prevent deep networks such as FCN~\cite{Long_2015_Fully} from being confused by the complicated and variable background contents,
\cite{Zhou_2017_Fixed} proposed to focus on a smaller input region according to an estimated bounding box.
On each viewpoint, two networks were trained for coarse-scaled segmentation and fine-scaled segmentation, respectively.
In the testing process, the coarse-scaled network was first used to obtain the rough position of the {\em pancreas},
and the fine-scaled network was executed several times and the segmentation mask was updated iteratively until convergence.

Despite the significant accuracy gain brought by this approach,
we notice a drawback originating from the {\em inconsistency} between its training and testing strategies.
That is to say, the training stage dealt with two networks individually without enabling global optimization,
but the testing phase assumed that they can cooperate with each other in a sequence of iterations.
From another perspective, a pixel-wise segmentation probability map was predicted by the coarse stage,
but the fine stage merely preserved the bounding box and discarded the remainder, which is a major information loss.
Sometimes, the image region within the bounding box does not contain sufficient spatial contexts,
and thus the fine stage can be confused and produce even lower segmentation accuracy than the coarse stage.
A failure case is shown in Figure~\ref{Fig:Motivation}.
This motivates us to connect these two stages with a saliency transformation module so as to jointly optimize their parameters.

\subsection{Recurrent Saliency Transformation Network}
\label{Approach:Formulation}

Following the baseline approach, we train an individual model for each of the three viewpoints.
Without loss of generality, we consider a 2D slice along the {\em axial} view, denoted by $\mathbf{X}_{\mathrm{A},l}$.
Our goal is to infer a binary segmentation mask $\mathbf{Z}_{\mathrm{A},l}$ of the same dimensionality.
In the context of deep neural networks~\cite{Long_2015_Fully}\cite{Chen_2015_Semantic},
this is often achieved by first computing a {\em probability map}
${\mathbf{P}_{\mathrm{A},l}}={\mathbf{f}\!\left[\mathbf{X}_{\mathrm{A},l};\boldsymbol{\theta}\right]}$,
where $\mathbf{f}\!\left[\cdot;\boldsymbol{\theta}\right]$ is a deep segmentation network
(FCN throughout this paper) with $\boldsymbol{\theta}$ being network parameters,
and then binarizing $\mathbf{P}_{\mathrm{A},l}$ into $\mathbf{Z}_{\mathrm{A},l}$ using a fixed threshold of $0.5$,
{\em i.e.}, ${\mathbf{Z}_{\mathrm{A},l}}={\mathbb{I}\!\left[\mathbf{P}_{\mathrm{A},l}\geqslant0.5\right]}$.

In order to assist segmentation with the probability map, we introduce $\mathbf{P}_{\mathrm{A},l}$ as a latent variable.
We introduce a {\em saliency transformation} module, which takes the probability map to generate an updated input image,
{\em i.e.}, ${\mathbf{I}_{\mathrm{A},l}}={\mathbf{X}_{\mathrm{A},l}\odot
    \mathbf{g}\!\left(\mathbf{P}_{\mathrm{A},l};\boldsymbol{\eta}\right)}$,
and uses the updated input $\mathbf{I}_{\mathrm{A},l}$ to replace $\mathbf{X}_{\mathrm{A},l}$.
Here $\mathbf{g}\!\left[\cdot;\boldsymbol{\eta}\right]$ is the transformation function with parameters $\boldsymbol{\eta}$,
and $\odot$ denotes element-wise product,
{\em i.e.}, the transformation function adds spatial weights to the original input image.
Thus, the segmentation process becomes:
\begin{equation}
\label{Eqn:RecurrentNetwork}
{\mathbf{P}_{\mathrm{A},l}}=
    {\mathbf{f}\!\left[\mathbf{X}_{\mathrm{A},l}\odot
    \mathbf{g}\!\left(\mathbf{P}_{\mathrm{A},l};\boldsymbol{\eta}\right);\boldsymbol{\theta}\right]}.
\end{equation}
This is a recurrent neural network.
Note that the saliency transformation function $\mathbf{g}\!\left[\cdot,\boldsymbol{\eta}\right]$
needs to be differentiable so that the entire recurrent network can be optimized in an end-to-end manner.
As $\mathbf{X}_{\mathrm{A},l}$ and $\mathbf{P}_{\mathrm{A},l}$ share the same spatial dimensionality,
we set $\mathbf{g}\!\left[\cdot,\boldsymbol{\eta}\right]$ to be a {\em size-preserved} convolution,
which allows the weight added to each pixel to be determined by the segmentation probabilities in a small neighborhood around it.
As we will show in the experimental section (see Figure~\ref{Fig:VisualizationNIH}),
the learned convolutional kernels are able to extract complementary information to help the next iteration.

\renewcommand{\figurewidth}{8.0cm}
\begin{figure}[t]
\begin{center}
    \includegraphics[width=\figurewidth]{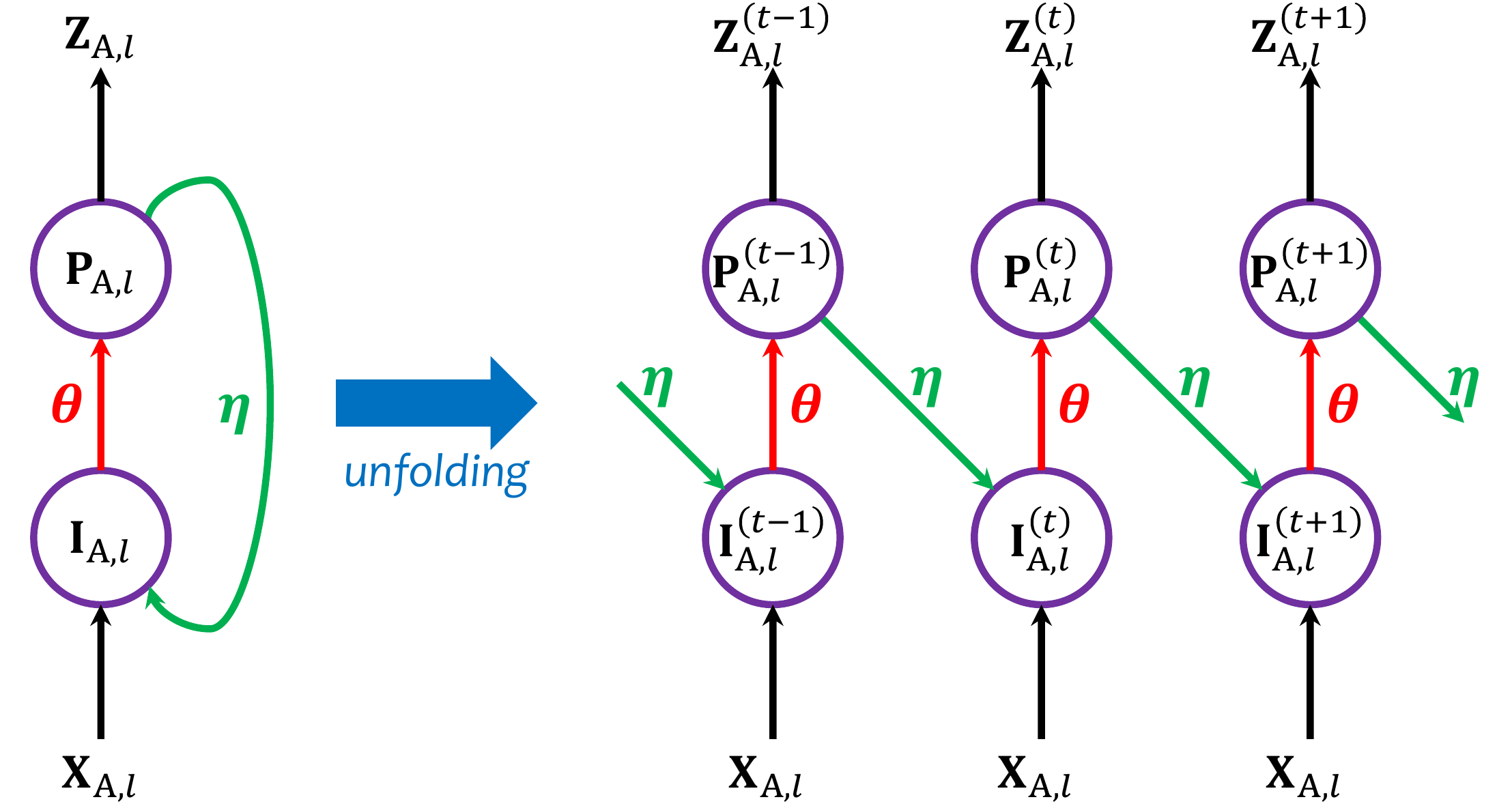}
\end{center}
\caption{
    We formulate our approach into a recurrent network, and unfold it for optimization and inference.
}
\label{Fig:RecurrentNetwork}
\end{figure}

\renewcommand{\figurewidth}{15.0cm}
\begin{figure*}[t]
\begin{center}
    \includegraphics[width=\figurewidth]{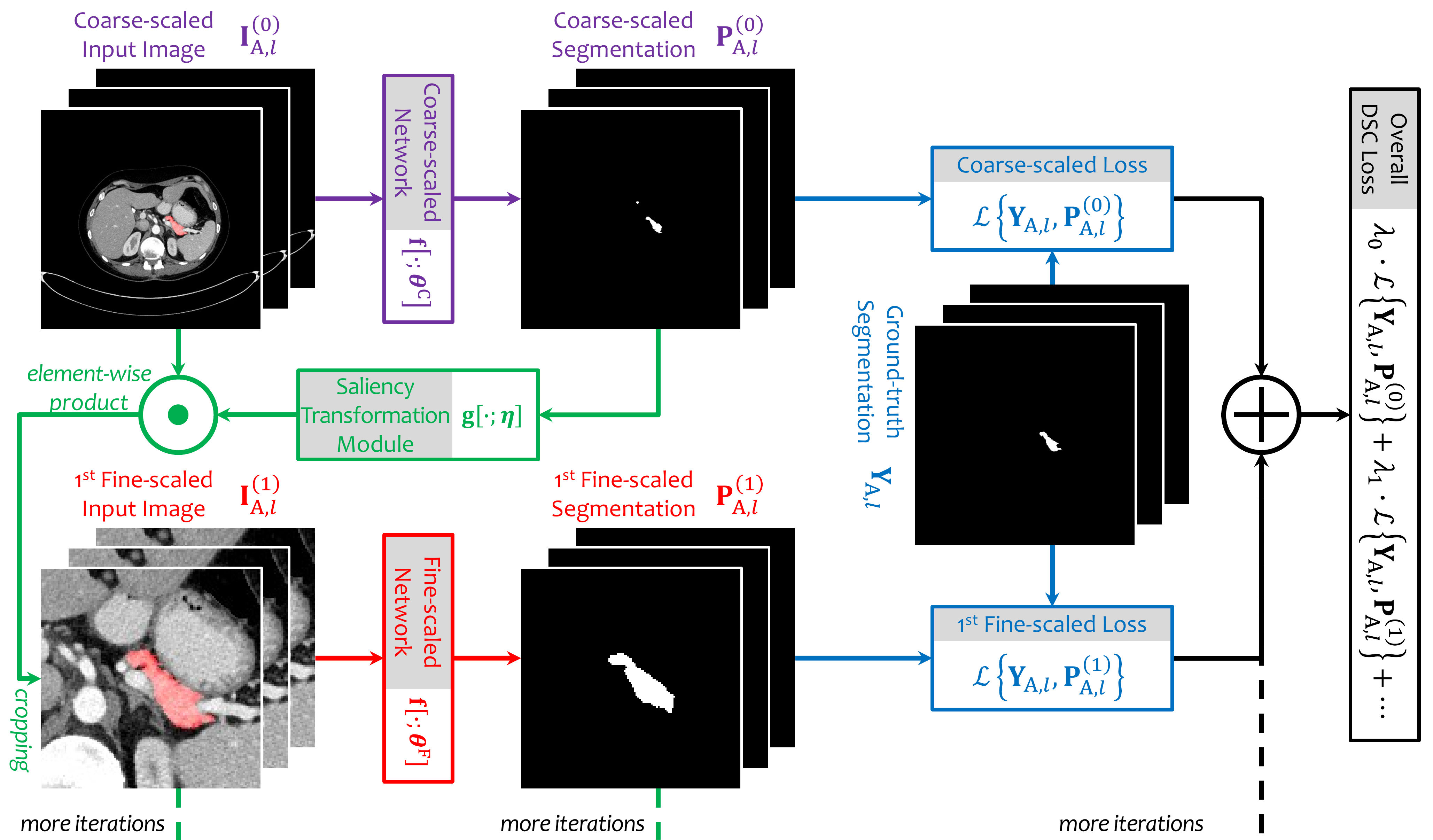}
\end{center}
\caption{
    Illustration of the training process (best viewed in color).
    We display an input image along the {\em axial} view which contains $3$ neighboring slices.
    To save space, we only plot the coarse stage and the first iteration in the fine stage.
}
\label{Fig:Framework}
\end{figure*}

To optimize Eqn~\eqref{Eqn:RecurrentNetwork},
we unfold the recurrent network into a plain form (see Figure~\ref{Fig:RecurrentNetwork}).
Given an input image $\mathbf{X}_{\mathrm{A},l}$ and an integer $T$ which is the maximal number of iterations,
we update $\mathbf{I}_{\mathrm{A},l}^{\left(t\right)}$ and $\mathbf{P}_{\mathrm{A},l}^{\left(t\right)}$, ${t}={0,1,\ldots,T}$:
\begin{eqnarray}
\label{Eqn:RecurrentComputation}
{\mathbf{I}_{\mathrm{A},l}^{\left(t\right)}} & = &
    {\mathbf{X}_{\mathrm{A},l}\odot\mathbf{g}\!\left(\mathbf{P}_{\mathrm{A},l}^{\left(t-1\right)};\boldsymbol{\eta}\right)}, \\
{\mathbf{P}_{\mathrm{A},l}^{\left(t\right)}} & = &
    {\mathbf{f}\!\left[\mathbf{I}_{\mathrm{A},l}^{\left(t\right)};\boldsymbol{\theta}\right]}.
\end{eqnarray}
Note that the original input image $\mathbf{X}_{\mathrm{A},l}$ does not change,
and the parameters $\boldsymbol{\theta}$ and $\boldsymbol{\eta}$ are shared by all iterations.
At ${t}={0}$, we directly set ${\mathbf{I}_{\mathrm{A},l}^{\left(0\right)}}={\mathbf{X}_{\mathrm{A},l}}$.

When segmentation masks $\mathbf{P}_{\mathrm{A},l}^{\left(t\right)}$ (${t}={0,1,\ldots,T-1}$) are available for reference,
deep networks benefit considerably from a shrunk input region especially when the target organ is very small.
Thus, we define a {\em cropping} function $\mathrm{Crop}\!\left[\cdot;\mathbf{P}_{\mathrm{A},l}^{\left(t\right)}\right]$,
which takes $\mathbf{P}_{\mathrm{A},l}^{\left(t\right)}$ as the {\em reference map},
binarizes it into ${\mathbf{Z}_{\mathrm{A},l}^{\left(t\right)}}=
    {\mathbb{I}\!\left[\mathbf{P}_{\mathrm{A},l}^{\left(t\right)}\geqslant0.5\right]}$,
finds the minimal rectangle covering all the activated pixels, and adds a $K$-pixel-wide margin (padding) around it.
We fix $K$ to be $20$; our algorithm is not sensitive to this parameter.

Finally note that $\mathbf{I}_{\mathrm{A},l}^{\left(0\right)}$, the original input (the entire 2D slice),
is much larger than the cropped inputs $\mathbf{I}_{\mathrm{A},l}^{\left(t\right)}$ for ${t}>{0}$.
We train two FCN's to deal with such a major difference in input data.
The first one is named the {\em coarse-scaled} segmentation network, which is used {\em only} in the first iteration.
The second one, the {\em fine-scaled} segmentation network, takes the charge of all the remaining iterations.
We denote their parameters by $\boldsymbol{\theta}^\mathrm{C}$ and $\boldsymbol{\theta}^\mathrm{F}$, respectively.
These two FCN's are optimized jointly.

We compute a DSC loss term on each probability map $\mathbf{P}_{\mathrm{A},l}^{\left(t\right)}$, ${t}={0,1,\ldots,T}$,
and denote it by $\mathcal{L}\!\left\{\mathbf{Y}_{\mathrm{A},l},\mathbf{P}_{\mathrm{A},l}^{\left(t\right)}\right\}$.
Here, $\mathbf{Y}_{\mathrm{A},l}$ is the ground-truth segmentation mask,
and ${\mathcal{L}\!\left\{\mathbf{Y},\mathbf{P}\right\}}={1-\frac{2\times{\sum_i}Y_iP_i}{{\sum_i}Y_i+P_i}}$
is based on a {\em soft} version of DSC~\cite{Milletari_2016_VNet}.
Our goal is to minimize the overall loss:
\begin{equation}
\label{Eqn:LossFunction}
{\mathcal{L}}={{\sum_{t=0}^T}\lambda_t\cdot
    \mathcal{L}\!\left\{\mathbf{Y}_{\mathrm{A},l}^{\left(t\right)},\mathbf{Z}_{\mathrm{A},l}^{\left(t\right)}\right\}}.
\end{equation}
This leads to joint optimization over all iterations, which involves network parameters $\boldsymbol{\theta}^\mathrm{C}$,
$\boldsymbol{\theta}^\mathrm{F}$, and transformation parameters $\boldsymbol{\eta}$.
$\left\{\lambda_t\right\}_{t=0}^T$ controls the tradeoff among all loss terms.
We set ${2\lambda_0}={\lambda_1}=\ldots={\lambda_T}={2/\left(2T+1\right)}$ so as to encourage accurate fine-scaled segmentation.

\subsection{Training and Testing}
\label{Approach:TrainingAndTesting}

\newcommand{\vkern}{-0.01cm}
\begin{algorithm}[t!]
\SetKwInOut{Input}{Input}
\SetKwInOut{Output}{Output}
\SetKwInOut{Return}{Return}
\Input{
    input volume $\mathbf{X}$, viewpoint ${\mathcal{V}}={\left\{\mathrm{C},\mathrm{S},\mathrm{A}\right\}}$;\\
\vspace{\vkern}
    parameters $\boldsymbol{\theta}_\mathrm{v}^\mathrm{C}$,
    $\boldsymbol{\theta}_\mathrm{v}^\mathrm{F}$ and $\boldsymbol{\eta}_\mathrm{v}$, ${\mathrm{v}}\in{\mathcal{V}}$;\\
\vspace{\vkern}
    max number of iterations $T$, threshold $\mathrm{thr}$;
}
\Output{
    segmentation volume $\mathbf{Z}$;
}
${t}\leftarrow{0}$, ${\mathbf{I}_\mathrm{v}^{\left(0\right)}}\leftarrow{\mathbf{X}}$, ${\mathrm{v}}\in{\mathcal{V}}$;\\
\vspace{\vkern}
${\mathbf{P}_{\mathrm{v},l}^{\left(0\right)}}\leftarrow
    {\mathbf{f}\!\left[\mathbf{I}_{\mathrm{v},l}^{\left(0\right)};\boldsymbol{\theta}_\mathrm{v}^\mathrm{C}\right]}$,
    ${\mathrm{v}}\in{\mathcal{V}}$, $\forall l$;\\
\vspace{\vkern}
${\mathbf{P}^{\left(0\right)}}={\frac{\mathbf{P}_\mathrm{C}^{\left(0\right)}+
    \mathbf{P}_\mathrm{S}^{\left(0\right)}+\mathbf{P}_\mathrm{A}^{\left(0\right)}}{3}}$,
    ${\mathbf{Z}^{\left(0\right)}}={\mathbb{I}\!\left[\mathbf{P}^{\left(0\right)}\geqslant0.5\right]}$;\\
\vspace{\vkern}
\Repeat{${t}={T}$\ {\bf or}\
    ${\mathrm{DSC}\!\left\{\mathbf{Z}^{\left(t-1\right)},\mathbf{Z}^{\left(t\right)}\right\}}\geqslant{\mathrm{thr}}$}{
    ${t}\leftarrow{t+1}$;\\
\vspace{\vkern}
    ${\mathbf{I}_{\mathrm{v},l}^{\left(t\right)}}\leftarrow
        {\mathbf{X}_{\mathrm{v},l}\odot\mathbf{g}\!\left(\mathbf{P}_{\mathrm{v},l}^{\left(t-1\right)};\boldsymbol{\eta}\right)}$,
        ${\mathrm{v}}\in{\mathcal{V}}$, $\forall l$;\\
\vspace{\vkern}
    ${\mathbf{P}_{\mathrm{v},l}^{\left(t\right)}}\leftarrow
        {\mathbf{f}\!\left[\mathrm{Crop}\!\left[
        \mathbf{I}_{\mathrm{v},l}^{\left(t\right)};\mathbf{P}_{\mathrm{v},l}^{\left(t-1\right)}\right];
        \boldsymbol{\theta}_\mathrm{v}^\mathrm{F}\right]}$, ${\mathrm{v}}\in{\mathcal{V}}$, $\forall l$;\\
\vspace{\vkern}
    ${\mathbf{P}^{\left(t\right)}}={\frac{\mathbf{P}_\mathrm{C}^{\left(t\right)}+
        \mathbf{P}_\mathrm{S}^{\left(t\right)}+\mathbf{P}_\mathrm{A}^{\left(t\right)}}{3}}$,
        ${\mathbf{Z}^{\left(t\right)}}={\mathbb{I}\!\left[\mathbf{P}^{\left(t\right)}\geqslant0.5\right]}$;\\
\vspace{\vkern}
}
\Return{
    ${\mathbf{Z}}\leftarrow{\mathbf{Z}^{\left(t\right)}}$.
}
\caption{
    The Testing Phase
}
\label{Alg:Testing}
\end{algorithm}

{\bf The training phase} is aimed at minimizing the loss function $\mathcal{L}$,
defined in Eqn~\eqref{Eqn:LossFunction}, which is differentiable with respect to all parameters.
In the early training stages, the coarse-scaled network cannot generate reasonable probability maps.
To prevent the fine-scaled network from being confused by inaccurate input regions,
we use the ground-truth mask $\mathbf{Y}_{\mathrm{A},l}$ as the reference map.
After a sufficient number of training,
we resume using $\mathbf{P}_{\mathrm{A},l}^{\left(t\right)}$ instead of $\mathbf{Y}_{\mathrm{A},l}$.
In Section~\ref{ExperimentsNIH:Settings}, we will see that this ``fine-tuning'' strategy improves segmentation accuracy considerably.

Due to the limitation in GPU memory, in each mini-batch containing one training sample,
we set $T$ to be the maximal integer (not larger than $5$) so that we can fit the entire framework into the GPU memory.
The overall framework is illustrated in Figure~\ref{Fig:Framework}.
As a side note, we find that setting ${T}\equiv{1}$ also produces high accuracy,
suggesting that major improvement is brought by joint optimization.

{\bf The testing phase} follows the flowchart described in Algorithm~\ref{Alg:Testing}.
There are two minor differences from the training phase.
First, as the ground-truth segmentation mask $\mathbf{Y}_{\mathrm{A},l}$ is not available,
the probability map $\mathbf{P}_{\mathrm{A},l}^{\left(t\right)}$ is always taken as the reference map for image cropping.
Second, the number of iterations is no longer limited by the GPU memory,
as the intermediate outputs can be discarded on the way.
In practice, we terminate our algorithm when the similarity of two consecutive predictions,
measured by ${\mathrm{DSC}\!\left\{\mathbf{Z}^{\left(t-1\right)},\mathbf{Z}^{\left(t\right)}\right\}}=
    {\frac{2\times{\sum_i}Z_i^{\left(t-1\right)}Z_i^{\left(t\right)}}
    {{\sum_i}Z_i^{\left(t-1\right)}+Z_i^{\left(t\right)}}}$,
reaches a threshold $\mathrm{thr}$, or a fixed number ($T$) of iterations are executed.
We will discuss these parameters in Section~\ref{ExperimentsNIH:Diagnosis:Convergence}.

\subsection{Discussions}
\label{Approach:Discussions}

Coarse-to-fine recognition is an effective idea in medical imaging analysis.
Examples include~\cite{Zhou_2017_Fixed}, our baseline, and~\cite{Chen_2016_Mitosis} for metosis detection.
Our approach can be applied to most of them towards higher recognition performance.

Attention-based or recurrent models are also widely used
for natural image segmentation~\cite{Chen_2016_Attention}\cite{Li_2017_Instance}\cite{Xia_2016_Zoom}\cite{Lin_2017_RefineNet}.
Our approach differs from them in making full use of the special properties of CT scans,
{\em e.g.}, each organ appears at a roughly fixed position, and has a fixed number of components.
Our approach can be applied to detecting the lesion areas of an organ~\cite{Kamnitsas_2017_Efficient}\cite{Zhou_2017_Deep},
or a specific type of vision problems such as {\em hair} segmentation in a {\em face}~\cite{Luo_2013_Structure},
or detecting the targets which are consistently small in the input images~\cite{Singh_2016_Learning}.

\section{Pancreas Segmentation Experiments}
\label{ExperimentsNIH}

\newcommand{\colwidthA}{2.2cm}
\newcommand{\colwidthB}{1.2cm}
\begin{table*}[!btp]
\centering
\begin{tabular}{|l||R{\colwidthA}|R{\colwidthB}||R{\colwidthB}|R{\colwidthB}|}
\hline
Model                                                             & Average         & Gain    & Max     & Min     \\
\hline\hline
Stage-wise segmentation~\cite{Zhou_2017_Fixed}                    & $82.37\pm 5.68$ & $    -$ & $90.85$ & $62.43$ \\
\hline\hline
Using $3\times3$ kernels in saliency transformation (basic model) & $83.47\pm 5.78$ & $+0.00$ & $90.63$ & $57.85$ \\
\hline
Using $1\times1$ kernels in saliency transformation               & $82.85\pm 6.68$ & $-0.62$ & $90.40$ & $53.44$ \\
\hline
Using $5\times5$ kernels in saliency transformation               & $83.64\pm 5.29$ & $+0.17$ & $90.35$ & $66.35$ \\
\hline\hline
Two-layer saliency transformation ($3\times3$ kernels)            & $83.93\pm 5.43$ & $+0.46$ & $90.52$ & $64.78$ \\
\hline\hline
Fine-tuning with noisy data ($3\times3$ kernels)                  & $83.99\pm 5.09$ & $+0.52$ & $90.57$ & $65.05$ \\
\hline
\end{tabular}
\caption{
    Accuracy (DSC, $\%$) comparison of different settings of our approach.
    Please see the texts in Section~\ref{ExperimentsNIH:Settings} for detailed descriptions of these variants.
    For each variant, the ``gain'' is obtained by comparing its accuracy with the basic model.
}
\label{Tab:Settings}
\end{table*}

\subsection{Dataset and Evaluation}
\label{ExperimentsNIH:DatasetAndEvaluation}

We evaluate our approach on the NIH {\em pancreas} segmentation dataset~\cite{Roth_2015_DeepOrgan},
which contains $82$ contrast-enhanced abdominal CT volumes.
The resolution of each scan is $512\times512\times L$,
where ${L}\in{\left[181,466\right]}$ is the number of slices along the long axis of the body.
The distance between neighboring voxels ranges from $0.5\mathrm{mm}$ to $1.0\mathrm{mm}$.

Following the standard cross-validation strategy, we split the dataset into $4$ fixed folds,
each of which contains approximately the same number of samples.
We apply cross validation, {\em i.e.}, training the models on $3$ out of $4$ subsets and testing them on the remaining one.
We measure the segmentation accuracy by computing the Dice-S{\o}rensen coefficient (DSC) for each sample,
and report the average and standard deviation over all $82$ cases.

\subsection{Different Settings}
\label{ExperimentsNIH:Settings}

We use the FCN-8s model~\cite{Long_2015_Fully} pre-trained on PascalVOC~\cite{Everingham_2010_Pascal}.
We initialize the up-sampling layers with random weights, set the learning rate to be $10^{-4}$ and run $80\rm{,}000$ iterations.
Different options are evaluated, including using different kernel sizes in saliency transformation,
and whether to fine-tune the models using the predicted segmentations as reference maps
(see the description in Section~\ref{Approach:TrainingAndTesting}).
Quantitative results are summarized in Table~\ref{Tab:Settings}.

As the saliency transformation module is implemented by a size-preserved convolution (see Section~\ref{Approach:Formulation}),
the size of convolutional kernels determines the range that a pixel can use to judge its saliency.
In general, a larger kernel size improves segmentation accuracy ($3\times3$ works significantly better than $1\times1$),
but we observe the marginal effect: the improvement of $5\times5$ over $3\times3$ is limited.
As we use $7\times7$ kernels, the segmentation accuracy is slightly lower than that of $5\times5$.
This may be caused by the larger number of parameters introduced to this module.
Another way of increasing the receptive field size is to use two convolutional layers with $3\times3$ kernels.
This strategy, while containing a smaller number of parameters, works even better than using one $5\times5$ layer.
But, we do not add more layers, as the performance saturates while computational costs increase.

As described in Section~\ref{Approach:TrainingAndTesting},
we fine-tune these models with images cropped from the coarse-scaled segmentation mask.
This is to adjust the models to the testing phase, in which the ground-truth mask is unknown,
so that the fine-scaled segmentation needs to start with, and be able to revise the coarse-scaled segmentation mask.
We use a smaller learning rate ($10^{-6}$) and run another $40\rm{,}000$ iterations.
This strategy not only reports $0.52\%$ overall accuracy gain,
but also alleviates over-fitting (see Section~\ref{ExperimentsNIH:Diagnosis:OverFitting}).

In summary, all these variants produce higher accuracy than the state-of-the-art ($82.37\%$ by~\cite{Zhou_2017_Fixed}),
which verifies that the major contribution comes from our recurrent framework which enables joint optimization.
In the later experiments, we inherit the best variant learned from this section,
including in a large-scale multi-organ dataset (see Section~\ref{ExperimentsMutliOrgan}).
That is to say, we use two $3\times3$ convolutional layers for saliency transformation,
and fine-tune the models with coarse-scaled segmentation.
This setting produces an average accuracy of $84.50\%$, as shown in Table~\ref{Tab:ComparisonNIH}.

\subsection{Comparison to the State-of-the-Art}
\label{ExperimentsNIH:Comparison}

\renewcommand{\colwidthA}{2.0cm}
\renewcommand{\colwidthB}{1.0cm}
\begin{table}[!btp]
\centering
\begin{tabular}{|l||R{\colwidthA}|R{\colwidthB}|R{\colwidthB}|}
\hline
Approach                                     & Average                  & Max              & Min              \\
\hline\hline
Roth {\em et al.}~\cite{Roth_2015_DeepOrgan} & $71.42\pm10.11$          & $86.29$          & $23.99$          \\
\hline
Roth {\em et al.}~\cite{Roth_2016_Spatial}   & $78.01\pm 8.20$          & $88.65$          & $34.11$          \\
\hline
Zhang {\em et al.}~\cite{Zhang_2016_Coarse}  & $77.89\pm 8.52$          & $89.17$          & $43.67$          \\
\hline
Roth {\em et al.}~\cite{Roth_2017_Spatial}   & $81.27\pm 6.27$          & $88.96$          & $50.69$          \\
\hline
Zhou {\em et al.}~\cite{Zhou_2017_Fixed}  & $82.37\pm 5.68$          & $90.85$          & $62.43$          \\
\hline
Cai {\em et al.}~\cite{Cai_2017_Improving}   & $82.4 \pm 6.7 $          & $90.1 $          & $60.0 $          \\
\hline\hline
Our Best Model                               & $\mathbf{84.50}\pm 4.97$ & $\mathbf{91.02}$ & $\mathbf{62.81}$ \\
\hline
\end{tabular}
\caption{
    Accuracy (DSC, $\%$) comparison between our approach and the state-of-the-arts
    on the NIH {\em pancreas} segmentation dataset~\cite{Roth_2015_DeepOrgan}.
    \cite{Zhang_2016_Coarse} was implemented in~\cite{Zhou_2017_Fixed}.
}
\label{Tab:ComparisonNIH}
\end{table}

We show that our approach works better than the baseline,
{\em i.e.}, the coarse-to-fine approach with two stages trained individually~\cite{Zhou_2017_Fixed}.
As shown in Table~\ref{Tab:ComparisonNIH}, the average improvement over $82$ cases is $2.13\pm2.67\%$,
which is impressive given such a high baseline accuracy ($82.37\%$ is already the state-of-the-art).
The standard deviations ($5.68\%$ of~\cite{Zhou_2017_Fixed} and $4.97\%$ of ours)
are mainly caused by the difference in scanning and labeling qualities.
The student's $t$-test suggests statistical significance (${p}={3.62\times10^{-8}}$).
A case-by-case study reveals that our approach reports higher accuracies on $67$ out of $82$ cases,
with the largest advantage being $+17.60\%$ and the largest deficit being merely $-3.85\%$.
We analyze the sources of improvement in Section~\ref{ExperimentsNIH:Diagnosis}.

Another related work is~\cite{Zhang_2016_Coarse} which stacks two FCN's for segmentation.
Our work differs from it by {\bf (i)} our model is recurrent, which allows fine-scaled segmentation to be updated iteratively,
and {\bf (ii)} we crop the input image to focus on the salient region.
Both strategies contribute significantly to segmentation accuracy.
Quantitatively, \cite{Zhang_2016_Coarse} reported an average accuracy of $77.89\%$.
Our approach achieves $78.23\%$ in the {\em coarse} stage, $82.73\%$ after {\em only one iteration},
and an entire testing phase reports $84.50\%$.

We briefly discuss the advantages and disadvantages of using 3D networks.
3D networks capture richer contextual information, but also require training more parameters.
Our 2D approach makes use of 3D contexts more efficiently.
At the end of each iteration, predictions from three views are fused,
and thus the saliency transformation module carries these information to the next iteration.
We implement VNet~\cite{Milletari_2016_VNet},
and obtain an average accuracy of $83.18\%$ with a 3D {\em ground-truth} bounding box provided for each case.
Without the ground-truth, a sliding-window process is required which is really slow --
an average of $5$ minutes on a Titan-X Pascal GPU.
In comparison, our approach needs $1.3$ minutes, slower than the baseline~\cite{Zhou_2017_Fixed} ($0.9$ minutes),
but faster than other 2D approaches~\cite{Roth_2015_DeepOrgan}\cite{Roth_2016_Spatial} ($2$--$3$ minutes).

\renewcommand{\figurewidth}{16.0cm}
\begin{figure*}[t]
\begin{center}
    \includegraphics[width=\figurewidth]{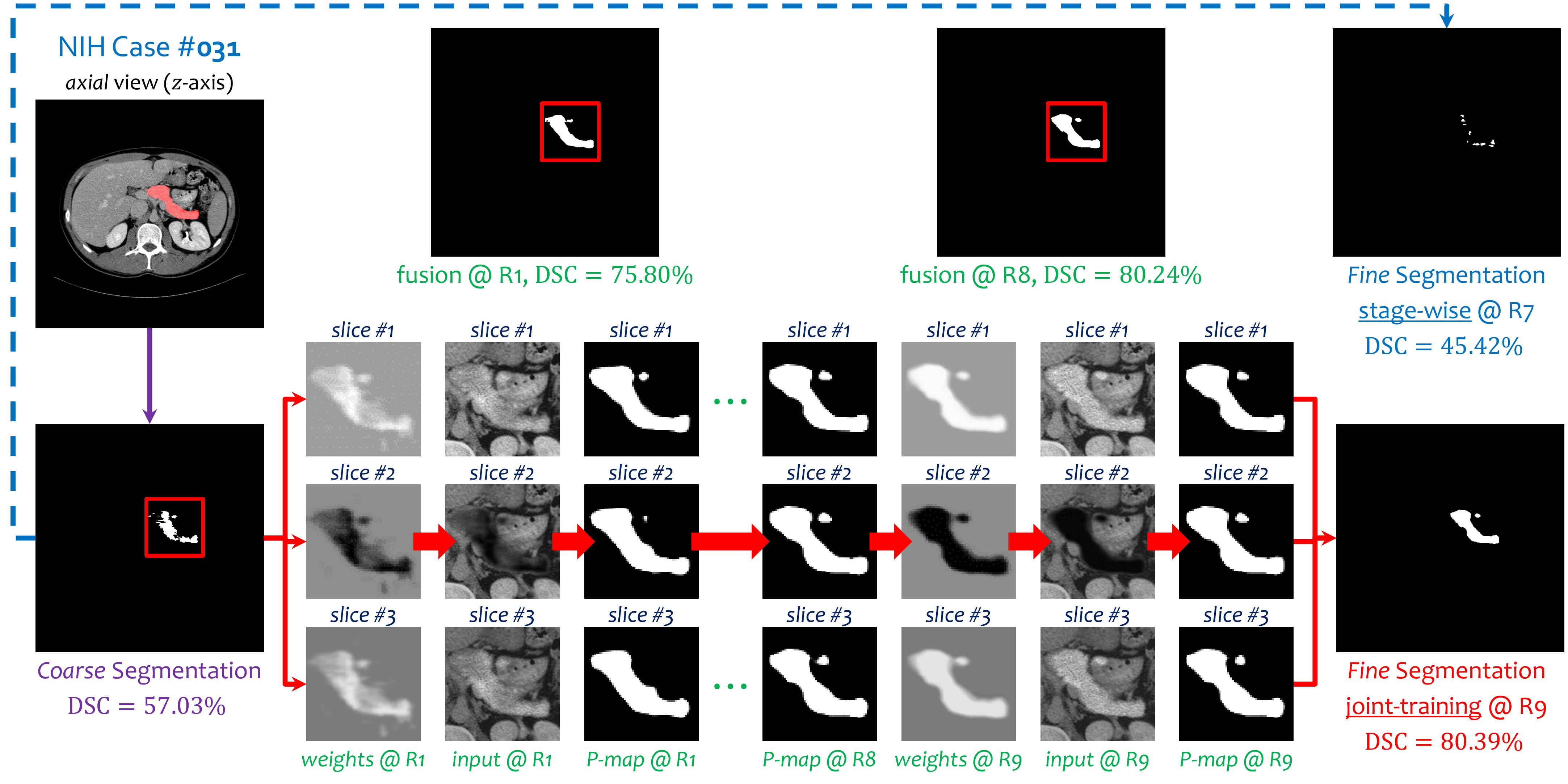}
\end{center}
\caption{
    Visualization of how recurrent saliency transformation works in coarse-to-fine segmentation (best viewed in color).
    This is a failure case of the stage-wise approach~\cite{Zhou_2017_Fixed} (see Figure~\ref{Fig:Motivation}),
    but segmentation accuracy is largely improved
    by making use of the probability map from the previous iteration to help the current iteration.
    Note that three weight maps capture different visual cues,
    with two of them focused on the foreground region, and the remaining one focused on the background region.
}
\label{Fig:VisualizationNIH}
\end{figure*}

\subsection{Diagnosis}
\label{ExperimentsNIH:Diagnosis}

\subsubsection{Joint Optimization and Mutli-Stage Cues}
\label{ExperimentsNIH:Diagnosis:JointOptimization}

Our approach enables joint training, which improves both the coarse and fine stages individually.
We denote the two networks trained in~\cite{Zhou_2017_Fixed} by $\mathbb{I}^\mathrm{C}$ and $\mathbb{I}^\mathrm{F}$,
and similarly, those trained in our approach by $\mathbb{J}^\mathrm{C}$ and $\mathbb{J}^\mathrm{F}$, respectively.
In the coarse stage, $\mathbb{I}^\mathrm{C}$ reports $75.74\%$ and $\mathbb{J}^\mathrm{C}$ reports $78.23\%$.
In the fine stage, applying $\mathbb{J}^\mathrm{F}$ on top of the output of $\mathbb{I}^\mathrm{C}$ gets $83.80\%$,
which is considerably higher than $82.37\%$ ($\mathbb{I}^\mathrm{F}$ on top of $\mathbb{I}^\mathrm{C}$)
but lower than $84.50\%$ ($\mathbb{J}^\mathrm{F}$ on top of $\mathbb{J}^\mathrm{C}$).
Therefore, we conclude that both the coarse-scaled and fine-scaled networks benefit from joint optimization.
A stronger coarse stage provides a better starting point, and a stronger fine stage improves the upper-bound.

In Figure~\ref{Fig:VisualizationNIH},
We visualize show how the recurrent network assists segmentation by incorporating multi-stage visual cues.
This is a failure case by the baseline approach~\cite{Zhou_2017_Fixed} (see Figure~\ref{Fig:Motivation}),
in which fine-scaled segmentation worked even worse because the missing contextual information.
It is interesting to see that in saliency transformation, different channels deliver complementary information,
{\em i.e.}, two of them focus on the target organ, and the remaining one adds most weights to the background region.
Similar phenomena happen in the models trained in different viewpoints and different folds.
This reveal that, except for foreground, background and boundary also contribute to visual recognition~\cite{Zhu_2017_Object}.

\vspace{-0.2cm}
\subsubsection{Convergence}
\label{ExperimentsNIH:Diagnosis:Convergence}

We study convergence, which is a very important criterion to judge the reliability of our approach.
We choose the best model reporting an average accuracy of $84.50\%$,
and record the inter-iteration DSC throughout the testing process: ${d^{\left(t\right)}}=
    {\mathrm{DSC}\!\left\{\mathbf{Z}^{\left(t-1\right)},\mathbf{Z}^{\left(t\right)}\right\}}=
    {\frac{2\times{\sum_i}Z_i^{\left(t-1\right)}Z_i^{\left(t\right)}}
    {{\sum_i}Z_i^{\left(t-1\right)}+Z_i^{\left(t\right)}}}$.

After $1$, $2$, $3$, $5$ and $10$ iterations, these numbers are $0.9037$, $0.9677$, $0.9814$, $0.9908$ and $0.9964$ for our approach,
and $0.8286$, $0.9477$, $0.9661$, $0.9743$ and $0.9774$ for~\cite{Zhou_2017_Fixed}, respectively.
Each number reported by our approach is considerably higher than that by the baseline.
The better convergence property provides us with the opportunity to set a more strict terminating condition,
{\em e.g.}, using ${\mathrm{thr}}={0.99}$ rather than ${\mathrm{thr}}={0.95}$.

We note that~\cite{Zhou_2017_Fixed} also tried to increase the threshold from $0.95$ to $0.99$,
but only $3$ out of $82$ cases converged after $10$ iterations, and the average accuracy went down from $82.37\%$ to $82.28\%$.
In contrary, when the threshold is increased from $0.95$ to $0.99$ in our approach,
$80$ out of $82$ cases converge (in an average of $5.22$ iterations),
and the average accuracy is improved from $83.93\%$ to $84.50\%$.
In addition, the average number of iterations needed to achieve ${\mathrm{thr}}={0.95}$
is also reduced from $2.89$ in~\cite{Zhou_2017_Fixed} to $2.02$ in our approach.
On a Titan-X Pascal GPU, one iteration takes $0.2$ minutes,
so using ${\mathrm{thr}}={0.99}$ requires an average of $1.3$ minutes in each testing case.
In comparison, \cite{Zhou_2017_Fixed} needs an average of $0.9$ minutes and~\cite{Roth_2016_Spatial} needs $2$-$3$ minutes.

\vspace{-0.2cm}
\subsubsection{The Over-Fitting Issue}
\label{ExperimentsNIH:Diagnosis:OverFitting}

Finally, we investigate the over-fitting issue by making use of {\em oracle} information in the testing process.
We follow~\cite{Zhou_2017_Fixed} to use the ground-truth bounding box {\em on each slice},
which is used to crop the input region in {\em every} iteration.
Note that annotating a bounding box in each slice is expensive and thus not applicable in real-world clinical applications.
This experiment is aimed at exploring the upper-bound of our segmentation networks under perfect localization.

With oracle information provided, our best model reports $86.37\%$,
which is considerably higher than the number ($84.50\%$) without using oracle information.
If we do not fine-tune the networks using coarse-scaled segmentation (see Table~\ref{Tab:Settings}),
the above numbers are $86.26\%$ and $83.68\%$, respectively.
This is to say, fine-tuning prevents our model from relying on the ground-truth mask.
It not only improves the average accuracy,
but also alleviates over-fitting (the disadvantage of our model against that with oracle information is decreased by $0.67\%$).

\section{Mutli-Organ Segmentation Experiments}
\label{ExperimentsMutliOrgan}

\renewcommand{\colwidthA}{1.1cm}
\begin{table}[!btp]
\centering
\begin{tabular}{|l||R{\colwidthA}|R{\colwidthA}||R{\colwidthA}|R{\colwidthA}|}
\hline
Organ               & \cite{Zhou_2017_Fixed}-{\bf C }  & \cite{Zhou_2017_Fixed}-{\bf F}
                    & Ours-{\bf C}                        & Ours-{\bf F}                        \\
\hline\hline
{\em adrenal g.}    & $57.38$          & $61.65$          & $60.70$          & $\mathbf{63.76}$ \\
\hline
{\em duodenum}      & $67.42$          & $69.39$          & $71.40$          & $\mathbf{73.42}$ \\
\hline
{\em gallbladder}   & $82.57$          & $^\sharp82.12$   & $87.08$          & $\mathbf{87.10}$ \\
\hline
{\em inferior v.c.} & $71.77$          & $^\sharp71.15$   & $79.12$          & $\mathbf{79.69}$ \\
\hline
{\em kidney l.}     & $92.56$          & $92.78$          & $96.08$          & $\mathbf{96.21}$ \\
\hline
{\em kidney r.}     & $94.98$          & $95.39$          & $95.80$          & $\mathbf{95.97}$ \\
\hline
{\em pancreas}      & $83.68$          & $85.79$          & $86.09$          & $\mathbf{87.60}$ \\
\hline
\end{tabular}
\caption{
    Comparison of coarse-scaled ({\bf C}) and fine-scaled ({\bf F}) segmentation
    by~\cite{Zhou_2017_Fixed} and our approach on our own dataset.
    A fine-scaled accuracy is indicated by $\sharp$ if it is lower than the coarse-scaled one.
    The {\em pancreas} segmentation accuracies are higher than those in Table~\ref{Tab:ComparisonNIH},
    due to the increased number of training samples and the higher resolution in CT scans.
}
\label{Tab:ComparisonMultiOrgan}
\end{table}

To verify that out approach applies to other organs,
we collect a large dataset which contains $200$ CT scans, $11$ abdominal organs and $5$ blood vessels.
This corpus took $4$ full-time radiologists around $3$ months to annotate.
To the best of our knowledge, this dataset is larger and contains more organs than any public datasets.
We choose $5$ most challenging targets including the {\em pancreas} and a blood vessel,
as well as two {\em kidneys} which are relatively easier.
Other easy organs such as the {\em liver} are ignored.
To the best of our knowledge, some of these organs were never investigated before,
but they are important in diagnosing pancreatic diseases and detecting the pancreatic cancer at an early stage.
We randomly partition the dataset into $4$ folds for cross validation.
Each organ is trained and tested individually.
When a pixel is predicted as more than one organs, we choose the one with the largest confidence score.

Results are summarized in Table~\ref{Tab:ComparisonMultiOrgan},
We first note that~\cite{Zhou_2017_Fixed} sometimes produced a lower accuracy in the fine stage than in the coarse stage.
Apparently this is caused by the unsatisfying convergence property in iterations,
but essentially, it is the loss of contextual information and the lack of globally optimized energy function.
Our approach solves this problem and reports a $4.29\%$ average improvement over $5$ challenging organs (the {\em kidneys} excluded).
For some organs, {\em e.g.}, the {\em gallbladder}, we do not observe significant accuracy gain by iterations.
But we emphasize that in these scenarios,
our coarse stage already provides much higher accuracy than the fine stage of~\cite{Zhou_2017_Fixed},
and the our fine stage preserves such high accuracy through iterations, demonstrating stability.
An example is displayed in Figure~\ref{Fig:VisualizationMultiOrgan}.

\renewcommand{\figurewidth}{8.0cm}
\begin{figure}[t]
\begin{center}
    \includegraphics[width=\figurewidth]{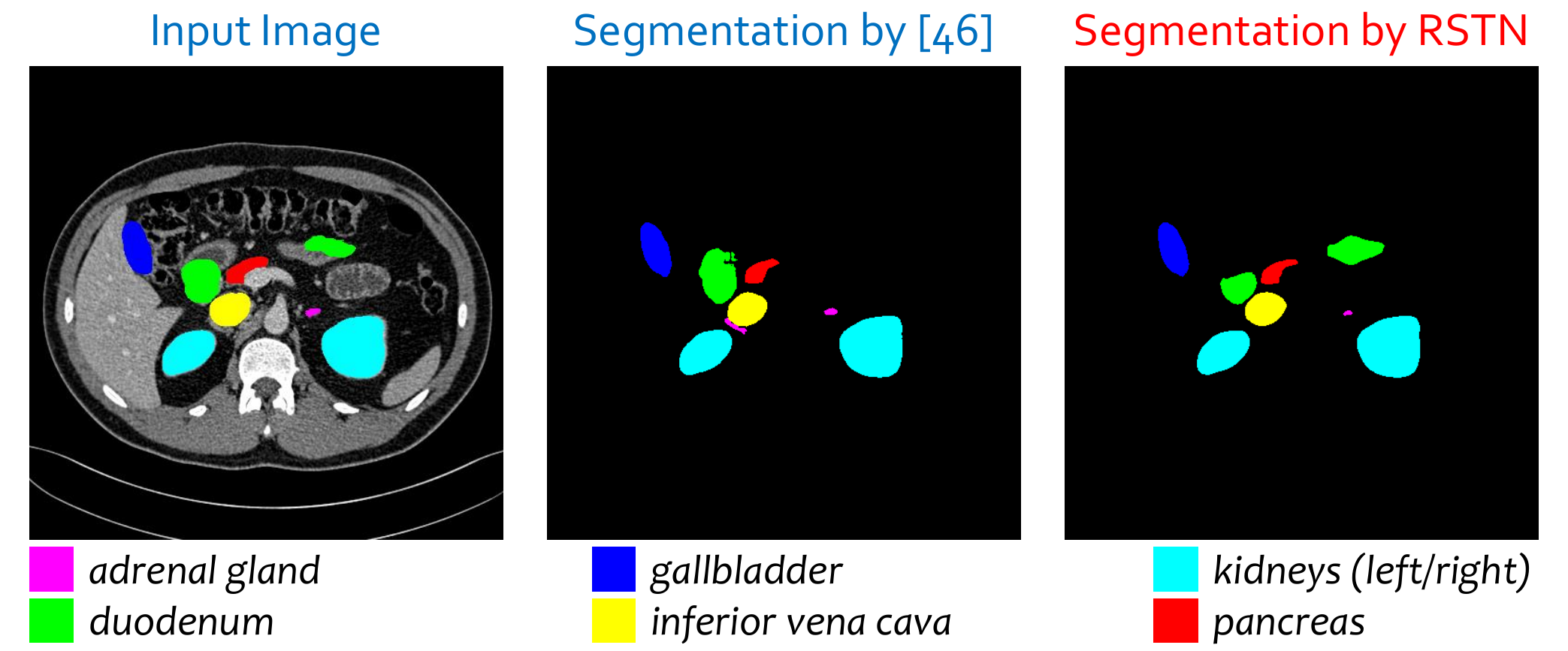}
\end{center}
\caption{
    Mutli-organ segmentation in the {\em axial} view (best viewed in color).
    Organs are marked in different colors (input image is shown with the ground-truth annotation).
}
\label{Fig:VisualizationMultiOrgan}
\end{figure}

\section{Conclusions}
\label{Conclusions}

This work is motivated by the difficulty of small organ segmentation.
As the target is often small, it is required to focus on a local input region,
but sometimes the network is confused due to the lack of contextual information.
We present the {\bf Recurrent Saliency Transformation Network}, which enjoys three advantages.
{\bf (i)} Benefited by a (recurrent) global energy function, it is easier to generalize our models from training data to testing data.
{\bf (ii)} With joint optimization over two networks, both of them get improved individually.
{\bf (iii)} By incorporating multi-stage visual cues, more accurate segmentation results are obtained.
As the fine stage is less likely to be confused by the lack of contexts, we also observe better convergence during iterations.

Our approach is applied to two datasets for {\em pancreas} segmentation and multi-organ segmentation,
and outperforms the baseline (the state-of-the-art) significantly.
Confirmed by the radiologists in our team, these segmentation results are helpful to computer-assisted clinical diagnoses.

\vspace{0.2cm}
\noindent
{\bf Acknowledgements:}
This paper was supported by the Lustgarten Foundation for Pancreatic Cancer Research.
We thank Wei Shen, Seyoun Park, Weichao Qiu,
Song Bai, Zhuotun Zhu, Chenxi Liu, Yan Wang, Siyuan Qiao, Yingda Xia and Fengze Liu for discussions.

{\small
\bibliographystyle{ieee}
\bibliography{egbib}
}

\end{document}